\documentclass[journal]{IEEEtran}
\IEEEoverridecommandlockouts
\usepackage{cite}
\usepackage{amsmath,amssymb,amsfonts}
\usepackage{graphicx}
\usepackage{float}
\usepackage{subfigure}
\usepackage{textcomp}
\usepackage{xcolor}
\usepackage{tabularx}
\usepackage{multirow}
\usepackage{algorithm}
\usepackage{algorithmicx}
\usepackage{algpseudocode}
\usepackage{soul}
\usepackage{verbatim}   

\def\BibTeX{{\rm B\kern-.05em{\sc i\kern-.025em b}\kern-.08em
    T\kern-.1667em\lower.7ex\hbox{E}\kern-.125emX}}

\begin{document}

\title{Boosting Gradient for White-Box Adversarial Attacks}
\author{Hongying~Liu,~\IEEEmembership{Member,~IEEE,}
        Zhenyu~Zhou,
        Fanhua~Shang,~\IEEEmembership{Senior Member,~IEEE,}
        Xiaoyu~Qi,
        Yuanyuan~Liu,~\IEEEmembership{Member,~IEEE,}
        and~Licheng~Jiao,~\IEEEmembership{Fellow,~IEEE}
\IEEEcompsocitemizethanks{\IEEEcompsocthanksitem H.\ Liu, Z.\ Zhou, F.\ Shang, X.\ Qi, Y.\ Liu and L.\ Jiao are with the Key Laboratory of Intelligent Perception and Image Understanding of Ministry of Education, Xidian University, China. Corresponding Author E-mail: \ fhshang@xidian.edu.cn.}
\thanks{Manuscript received October 9, 2020.}
}

\maketitle

\begin{abstract}
Deep neural networks (DNNs) are playing key roles in various artificial intelligence applications such as image classification and object recognition. However, a growing number of studies have shown that there exist adversarial examples in DNNs, which are almost imperceptibly different from original samples, but can greatly change the network output. Existing white-box attack algorithms can generate powerful adversarial examples. Nevertheless, most of the algorithms concentrate on how to iteratively make the best use of gradients to improve adversarial performance. In contrast, in this paper, we focus on the properties of the widely-used ReLU activation function, and discover that there exist two phenomena (i.e., wrong blocking and over transmission) misleading the calculation of gradients in ReLU during the backpropagation. Both issues enlarge the difference between the predicted changes of the loss function from gradient and corresponding actual changes, and mislead the gradients which results in larger perturbations. Therefore, we propose a universal adversarial example generation method, called ADV-ReLU, to enhance the performance of gradient based white-box attack algorithms. During the backpropagation of the network, our approach calculates the gradient of the loss function versus network input, maps the values to scores, and selects a part of them to update the misleading gradients. Comprehensive experimental results on \emph{ImageNet} demonstrate that our ADV-ReLU can be easily integrated into many state-of-the-art gradient-based white-box attack algorithms, as well as transferred to black-box attack attackers, to further decrease perturbations in the ${\ell _2}$-norm.
\end{abstract}



\section{Introduction}
\IEEEPARstart{A}{s} an important branch in artificial intelligence, deep neural networks have made remarkable achievements in various fields. However, a deep neural network usually contains deep structure, complex learning parameters and massive training data, which makes it difficult for human to intuitively understand. With the discovery of adversarial examples in image classification networks \cite{szegedy2013intriguing,goodfellow2014explaining}, researchers are increasingly concerned about how to generate adversarial examples efficiently \cite{kurakin2016adversarial,Carlini2016Towards,MoosaviDezfooli2016DeepFoolAS}, improve the transferability of adversarial examples \cite{dong2018boosting,wu2018understanding,moosavi2017universal,Shi2019CurlsW}, promote the robustness of the network \cite{goodfellow2014explaining,kurakin2016adversarial,Tram2017Ensemble,Xie2019FeatureDF,AdSurvey19}, and explain the existence of adversarial examples \cite{Ilyas2019Adversarial} in deep neural networks.

In the methods of generating adversarial examples, according to the known information of the target network, they can be divided into two categories \cite{AdSurvey19, AdSurvey20, tnnlsGrad}: white-box attacks and black-box attacks. The difference between them is that white-box attackers know more information in the target network structure than black-box ones. For example, in the white-box attacks, the attacker can access the gradients, structures, parameters and even training data of the target network, while in the black-box attacks, the attacker does not have any knowledge about the target network. Compared with black-box attacks, white-box attacks can generate smaller perturbation and generally take less time than black-box attacks. And the white-box attacks are often used to generate adversarial training samples for adversarial training or to thoroughly evaluate the robustness of the network. Therefore, this paper focuses on improvement schemes for white-box attacks.

\begin{figure}[t]  
\centering
\includegraphics[width=0.47\textwidth, height=0.35\textwidth]{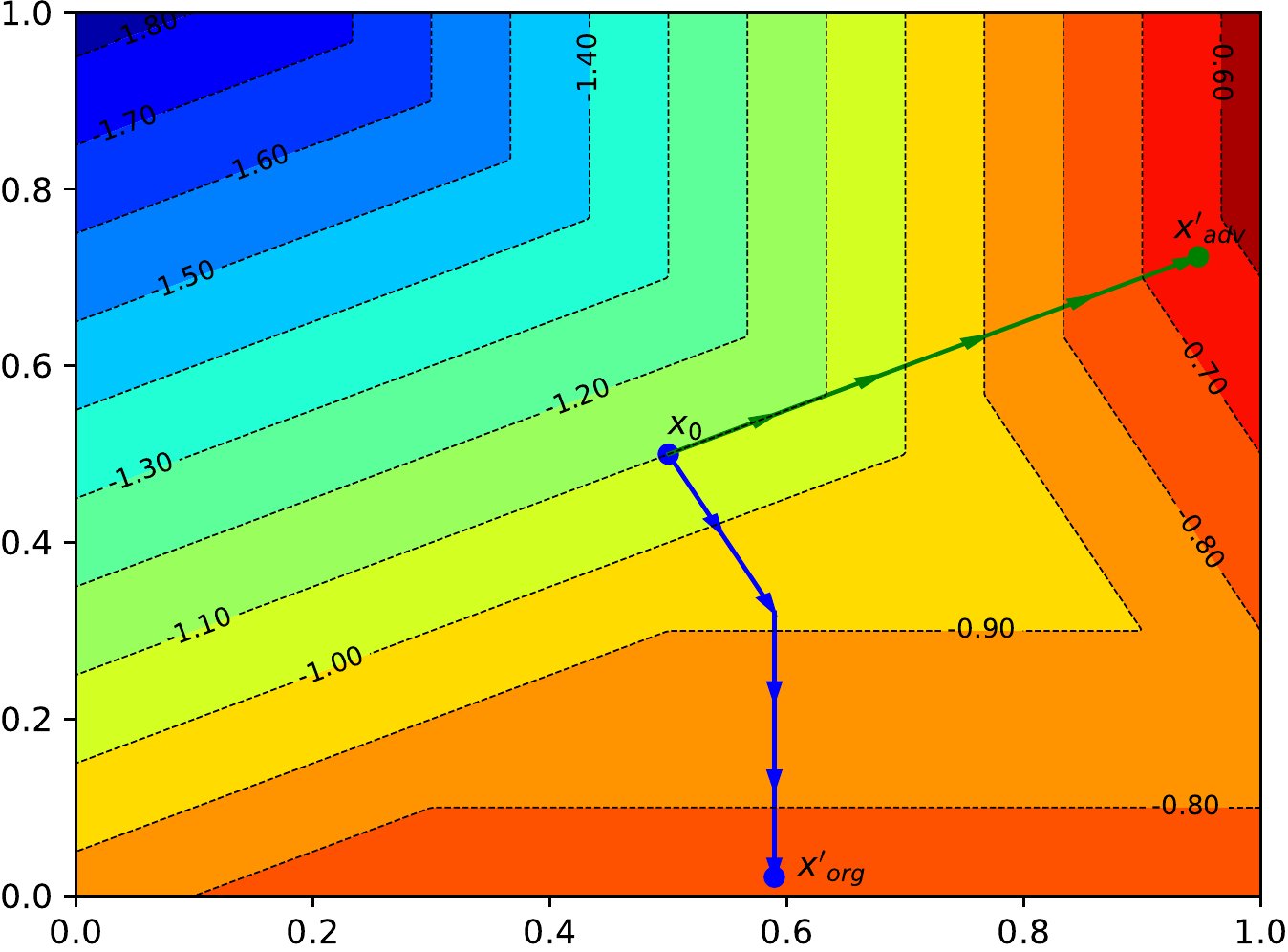}
\caption{Iterative trajectories of two types of gradient descents using the same step size. Note that $x_0 \rightarrow x'_{org}$ (blue) is with the original ReLU, while $x_0 \rightarrow x'_{adv}$ (green) is with the proposed ADV-ReLU.}
\label{Fig.contour}	
\end{figure}

For the white-box attacks, one of the major categories are the gradient-based attack methods. They usually construct a loss function that can cause the improvement of the perturbation attacking ability and decreasing of the perturbation magnitude, then optimizes the loss function through gradients to generate adversarial examples, such as L-BFGS \cite{szegedy2013intriguing} and C\&W \cite{Carlini2016Towards}. To generate adversarial examples under multiple constraints, L-BFGS proposes a box constrained L-BFGS to solve them approximatively. In C\&W, \cite{Carlini2016Towards} improves L-BFGS by inputting field mapping, changing loss function and using adaptive parameter search. Although it takes much time to search parameters adaptively, the algorithm can further reduce perturbations to increase the robustness of the network after adversarial training. There are also a series of  Fast Gradient Sign Method (FGSM) algorithms. Goodfellow \emph{et al}.\ \cite{goodfellow2014explaining} first proposed the FGSM algorithm. The FGSM algorithm takes a sign function to the gradient which is the network loss function versus input data, then multiplies the resulting value with a fixed coefficient to obtain the final noise. Thus, the algorithm can effectively control the noise level in the ${\ell_\infty}$ norm by controlling the coefficient. In order to improve the effect and reduce the perturbation, Kurakin \emph{et al}.\ \cite{kurakin2016adversarial} proposed the I-FGSM algorithm. Compared with FGSM, this algorithm adopted a better iterative optimizer to generate adversarial examples. In this way, I-FGSM finds much smaller perturbations than FGSM. Moreover, MI-FGSM \cite{dong2018boosting} and  vr-IGSM \cite{wu2018understanding} have also been presented. They gain effective transferability of perturbations by smoothing the iteration process during the generation of adversarial
examples.

However, in the white-box attack case, all of the above algorithms require the gradient of the target network, and they ignore the role played by the activation function during back propagation to accumulate gradient, and fail to optimize the gradient itself. During the analysis of gradient-based white-box attack algorithms, there is one issue, which can be easily ignored is that these powerful white-box attack algorithms focus on how to make better use of gradients to search adversarial examples, while ignoring the process of how deep networks attain the gradient of the loss function. We find a fact that there exist two phenomena, which mislead the calculation of gradients during back propagation passing through the ReLU function, that are named as ``wrong blocking" and ``over-transmission", respectively. These two phenomena indicate that, during the backpropagation, ReLU wrongly blocks and overly transmits gradient messages on some neurons. The two defects from the ReLU derivation processes mislead the backpropagation process, enlarge the difference between the expectation of gradient and the actual change as shown in Fig. \ref{Fig.contour}, and thus lower the quality of generated adversarial examples.

To address these two issues, we propose a general universal adversarial example generation method, called ADV-ReLU, that can be wildly used in existing gradient-based white-box attack algorithms and improve their effectiveness. In our framework, the attack algorithm can be mainly divided into two parts: ADV-ReLU-B (block) for improving the ``wrong blocking" and ADV-ReLU-T (transmit) for alleviating the ``over-transmission",  respectively. The key idea of ADV-ReLU-B is that, during the derivation process of ReLU, the elements in its input feature maps are sorted according to the severity of ``wrong blocking", then select a small part from these ranked elements as the most needed continue backpropagation elements, at which we modify the derivative rule of ReLU to renew the gradients. Similarly, ADV-ReLU-T sorts the elements of RELU's input feature maps by the severity of  ``over-transmission", then selects a small number of most  ``over transmission" elements and stop them propagating. Finally, we combine the two algorithms to obtain ADV-ReLU, which can prevent ReLU from both ``wrong blocking" and  ``over transmission". Our ADV-ReLU algorithm can easily be integrated with various existing gradient-based white-box attack algorithms, and thus resulting white-box attack algorithms can take the advantage of more accurate gradients to further increase the attack capability. Therefore, we can utilize the proposed more powerful attack algorithm to obtain more robust deep networks for various real-word applications.

Our main contributions are summarized as follows:
\begin{enumerate}
    \item During the backpropagation process in deep neural networks, we find two defects that are ignored by conventional works, when passing by the widely used ReLU activation function. That is, ``wrong blocking" and ``over transmission" cause a misleading to the gradient computation, and decrease consistency between the expectation of gradients and the actual change.
    \item We propose a universal adversarial example generation framework (ADV-ReLU) for gradient-based white-box attack algorithms. This framework can address the issues of ``wrong blocking" and ``over transmission", and it can be easily integrated into gradient-based white-box attack algorithms for further improving their performance.
    \item We conduct extensive experiments on ImageNet, demonstrating that, under many powerful white-box attacks, our framework obtains smaller ${\ell _2}$ norm perturbations in the attack results. Moreover, we also indicate the strategies to improve the robustness of networks by adversarial training.
\end{enumerate}

The rest of this paper is organized as follows. In Section II, the background of gradient-based attacks is briefly introduced. Section III presents our general universal adversarial example generation method. Section IV shows the experimental results and analysis. In Section V, we conclude our work.

\section{Background}
In this section, we introduce some related work about adversarial example methods.

We describe the process of non-targeted generating adversarial examples for image classification as follows:
\begin{gather}
    \min_{x'} \|x'-x\|_p \notag \\
    \textup{s.t.,}\;\; f(x')\neq f(x) \\
    x'\in [0,1] \notag
\end{gather}
where $x$ denotes the original input data, and $x'$ is an adversarial example. Note that $\|\cdot\|_p:\mathbb{R}^n\rightarrow\mathbb{R}$ represents the $\ell_p$-norm, where usually $p=0, 1, 2$ or $\infty$, which is used to evaluate the distance between input and adversarial samples. $f(\cdot):\mathbb{R}^n\rightarrow\mathbb{N}$ denotes the deep neural network for classification, which directly outputs the image categories. In addition, $l\!=\!f(x)$ is the original category of sample $x$, which also means the network can classify it correctly. $x'\in[0,1]$ indicates that the pixel value of the generated adversarial example still belongs to the predetermined image range (i.e., an integer in range of $[0,255]$). In order to achieve the purpose of misclassifying the image (i.e., $f(x')\neq f(x)$), some adversarial attack algorithms construct a loss function $J_f (x)$ to evaluate the degree of misclassification of the network, then optimize $J_f (x)$ by changing input data $x$. During the process of $x\rightarrow x'$, when $f(x')$ is not equal to $f(x)$, we can judge that the adversarial example is generated successfully.

In a white-box attack, the attackers can easily obtain the model architectures and weights, and calculate the gradient from the input data versus the loss function: $\nabla\! J_f (x)$, to optimize the loss function. In contrast, the black-box attack can not access the model parameters. Next, we briefly introduce several white-box and black-box attack algorithms.

\subsection{Fast Gradient Sign Method (FGSM)}
The Fast Gradient Sign Method (FGSM) proposed in \cite{goodfellow2014explaining} generally sets the loss function as cross entropy on which the target network is trained, and then increases the perturbation in the direction that one-step toward the sign function of the loss function gradient, so that the algorithm can easily control the perturbation magnitude in the $\ell_\infty$-norm.
\begin{equation}
    x' = x + \alpha  \cdot \textrm{sign}(\nabla {J_f}(x)).
\end{equation}

\subsection{Iterative FGSM (I-FGSM)}
In \cite{kurakin2016adversarial}, the iterative FGSM (I-FGSM) was proposed. Obviously, in the neighborhood of the input data $x$, the gradient of the loss function is constantly changing. Moreover, I-FGSM can change the direction of the generated perturbation according to the current optimal gradient at each iteration, thus generating a simple but very powerful adversarial attack.
\begin{equation}
    x_{t+1}^{\prime}=\textrm{Clip}_{x, \varepsilon}\left\{x_{t}^{\prime}+\alpha \cdot \textrm{sign}\left(\nabla J_{f}(x)\right)\right\}
\end{equation}
where Clip denotes a clipping function, and $\operatorname{Clip}_{x, \varepsilon}$ means that the obtained adversarial example is
re-mapped back into the norm constraint $\varepsilon$, so as to control the size of the
final perturbation.

\subsection{Curls \& Whey}
The Curls\&Whey method \cite{Shi2019CurlsW} is a black-box attack, and can be divided into two parts, Curls iteration and Whey
optimization. The ordinary I-FGSM optimizing path can be regarded as a single gradient ascent path on cross entropy. However, during the Curls iteration process, we add an additional path which is first gradient descent, then ascent, and finally select the smallest perturbation among the two paths. In this way, because of the increasing of diversity of generating path, this method greatly improves the success rate for the adversarial example transferred to other networks. In addition, in the part of Whey optimization, Curls\&Whey further ``squeezes" the adversarial examples as close as possible to the classification boundary, reducing the excessive perturbation.

Moreover, since the black-box attacks can not access the model parameters of the attacked network, the substitute model is proposed to study the transferability \cite{Papernot2016TransferabilityIM} of the white-box attacks. That is, the attacker uses the white-box algorithms with the parameters, e.g., the gradient from a substitute model to attack the target networks.

Several works try to explain why adversarial examples exist. For example, \cite{goodfellow2014explaining} regards the appearance of adversarial as the linearization across the networks. And according to a series of experiments of adversarial attacks to robust or non-robust networks, \cite{Ilyas2019Adversarial} interprets adversarial examples as the human-imperceptible features, which are benefit to facilitating the classification correctly during the normal training for a network. Different from the above works, we focus on the backpropagation process and the role of the activation function in a neural network.

\section{Methodology}
In this section, we elaborate our framework, ADV-ReLU, based on gradient in details. The main idea of this method is that by changing the
derivation of ReLU during the backward prorogation of the deep neural networks, we can alleviate two misleading phenomena to obtain the gradient with higher guiding significance, therefore, promoting the performance of attack algorithms. Since our algorithm is based on the backward propagation, it can be embedded in other gradient-based white-box attacks.
In this section, we first analyze the backward process during generating adversarial examples to discover two unreasonable aspects at the derivation of the ReLU function. We then propose two algorithms: ADV-ReLU-B and ADV-ReLU-T to alleviate the two defects, respectively. Finally, we integrate these algorithms into ADV-ReLU, and show its property.

\begin{figure*}[t]
	\centering  
	\includegraphics[width=0.7\textwidth]{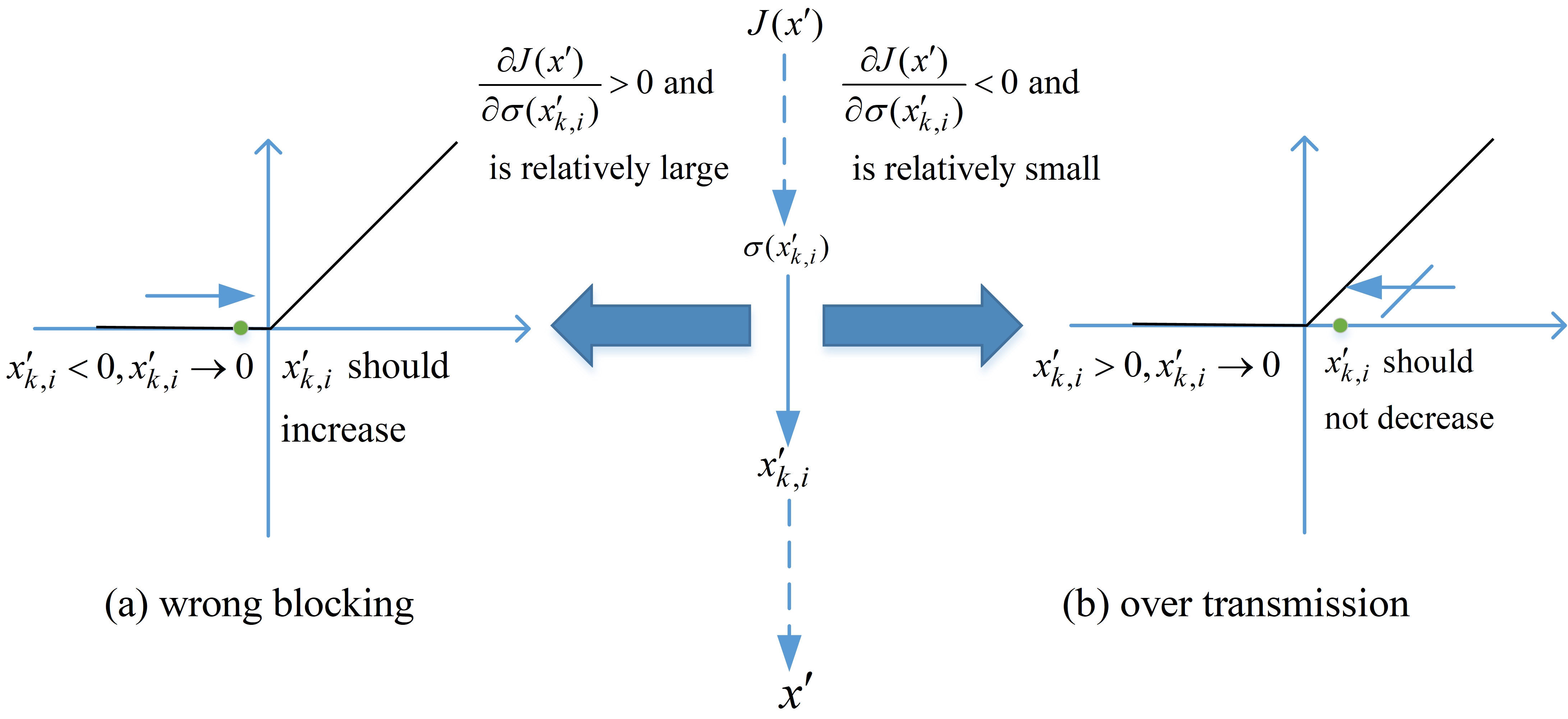}
	\caption{A diagram of ``wrong blocking" and ``over transmission". The middle arrows pointing down represent the process of backpropagation in a network, and the green point in the left part of (a) is $x'_{k,i}$, and the small arrow nearby it represents the direction that we want to change. Similarly, the right part of (b) represents ``over transmission", where the stopping small arrow represents the direction that we do not want to change.}
    \label{Fig.adv relu two drawbacks}	
\end{figure*}


\subsection{Drawbacks of backpropagation with ReLU}
\label{TwoDrawbacks}
During the generation of an adversarial example $x^{\prime}$, the calculation process of a loss function $J(x^\prime)$ in deep neural networks with the ReLU activation function can be usually formulated as follows:
\begin{equation}
J(x^\prime)={loss}\left(f\left(x^{\prime}\right), \;\text{label}\right)
\end{equation}
where ${loss}(\cdot)$ denotes the common loss function such as cross-entropy, and $f(\cdot)$ represents a deep neural network with $L\in\mathbb{Z}$ layers, which outputs the prediction of input $x^{\prime}$. $\sigma(\cdot)$ is an activation function and its input is $x_k^{\prime}$, which is a network feature map at the $k$-th layer, where $k\in[2,L]$. Note that for convenience, the activation function of all the layers is denoted by $\sigma(\cdot)$. In fact, the last layer usually uses the softmax function, and the remaining ones use the ReLU activation function. Therefore, the derivative path of the loss $J(x^\prime)$ versus input $x^{\prime}$ only passing through the ReLU can be represented as:
\begin{equation}
\frac{\partial J(x^\prime)}{\partial x^{\prime}}=\frac{\partial J(x^\prime)}{\partial \sigma(x_{L}^{\prime})}\frac{\partial \sigma(x_{L}^{\prime})}{\partial x_{L-1}^{\prime}} \cdots\cdot\frac{\partial \sigma(x^{\prime})}{\partial x^{\prime}}.
\end{equation}


In most cases, we can simply convert a minimizing problem into an equivalent negative maximizing one. Suppose that we should optimize $x'$ in the direction of increase the loss function $J(x^\prime)$, which equals to $\mathop {\max }\limits_{x'} {\rm{ }}J(x')$. The expression $x_{k,i}^\prime$ denotes the $i$-th element of the feature map $x_{k}^\prime$, where $i\in [1,M_k]$, and $M_k$ is the number of the elements in $x_{k}^\prime$.

As shown in Fig.\ \ref{Fig.adv relu two drawbacks}, the two drawbacks from the ReLU derivation processes are listed as follows.
\begin{itemize}
\item When $\frac{\partial J(x^\prime)}{\partial \sigma\left(x_{k,i}^{\prime}\right)} > 0$ and is relatively large, and $x_{k,i}^{\prime} \leq 0$ and is very close to 0, as shown in Fig.\ \ref{Fig.adv relu two drawbacks}(a), the increasing of $x_{k,i}'$ plays a major role for $\max {\rm{ }}{J}(x')$ because of the condition of $\frac{\partial J(x^\prime)}{\partial \sigma\left(x_{k,i}^{\prime}\right)}$. However, due to the property of the derivative of ReLU that $x_{k,i}^{\prime}<0 \rightarrow \frac{\partial \sigma\left(x_{k,i}^{\prime}\right)}{\partial x_{k,i}^{\prime}}=0 \rightarrow \frac{\partial J(x^\prime)}{\partial x_{k,i}^{\prime}}=0$, it results in the wrongly block in the backpropagation chain passing by ReLU, so that it blocks valuable information. We call this phenomenon as ``wrong blocking".
\item When $\frac{\partial J(x^\prime)}{\partial \sigma\left(x_{k,i}^{\prime}\right)} < 0$ is relatively small, and $x_{k,i}^{\prime} > 0$ and is very close to 0, as shown in Fig.\ \ref{Fig.adv relu two drawbacks}(b), $x_{k,i}^{\prime}>0 \rightarrow \frac{\partial\sigma\left(x_{k,i}^{\prime}\right)}{\partial x_{k,i}^{\prime}}=1$. Suppose we are to significantly decrease $x'_{k,i}$ according to the inherited gradient $\frac{\partial J(x^\prime)}{\partial \sigma\left(x_{k,i}^{\prime}\right)}$ with large magnitude, then $x'_k$ soon will move toward the negative axis redundantly to stop backpropagation for $x_{k}^{\prime}$ is so close to 0. The gradient is overly transferred, and we call this drawback as ``over transmission".
\end{itemize}

\subsection{Improvement for ``wrong blocking"}
In order to alleviate or even eliminate the issue of ``wrong blocking", we should not simply change ReLU into other activation functions maintaining continuous derivation, such as ELU \cite{Clevert2015FastAA} and Leaky-ReLU \cite{Maas13rectifiernonlinearities}. The reason is that this will produce lots of gradients mismatching the original gradients, thereby extremely perturbing the backward process. One appropriate solution is only to choose a small number of valuable elements blocked by ``wrong blocking", then change the derivative rule at these elements to let them pass the gradient information successfully. In this way, we can not only avoid excessive perturbation to normal backward for changing the derivative of ReLU, but also hold the important backward route unblocked to gain more precise gradient information. To satisfy these conditions, we propose a new rank-based algorithm to improve this defect, and name it as ADV-ReLU-B, as listed in Algorithm 1.

\begin{algorithm}[t] 
\caption{ADV-ReLU-B} 
\label{alg.ADV-ReLU-before} 
\renewcommand{\algorithmicrequire}{\textbf{Input:}}  
\renewcommand{\algorithmicensure}{\textbf{Output:}} 
\begin{algorithmic}[1] 
\Require Input feature maps $u$ for ReLU, inherited gradient $g_{b}=\frac{\partial J(u)}{\partial \sigma(u)}$, sort rate $s$, and a constant $c$.
\Ensure Modified gradient $g=\frac{\partial{J(u)}}{\partial u}$.
\State Select and obtain the set $I_B = \{ i | u[i]\le0,{g_b}[i]>0\}$;\;\;\;\quad \hfill $/\!/$\,\emph{$u[i]$ represents the $i$-th element of the feature maps $u$, and $u$ has the same size as $g_{b}$}
\State Calculate the set $V = \left\{ {c \cdot u[i] + {g_b}[i]|i \in I_B} \right\}$; \!\!\hfill $/\!/$\,\emph{$V$ collects the values of backpropagation of the $I_B$ elements}
\State Sort $V$ in a descending order;
\State Select $\lfloor s \cdot|V|\rfloor$ largest elements in $V$ to obtain $G$, where the number is $\left| G \right| = \lfloor s \cdot|V|\rfloor$;
\State Let $g_{\sigma}=\frac{\partial \sigma(u)}{\partial u}$, then cover the elements at the index set $G$ to renew the gradients, i.e., $g_{\sigma}[G]=1$;
\State $g=g_{b} \odot g_{\sigma}$; \!\!\hfill $/\!/$\,\emph{$\odot$ is element-wise multiplication}
\State return  $g$.
\end{algorithmic}
\end{algorithm}

In fact, our modified ReLU activation function used in Algorithm 1 does not change the property of forward process, while it changes the process of backward slightly (i.e., a small sort rate $0<s<1$ controls the proportion of selecting). Similar to most of the backward algorithms, it requires the backward inherited gradient $g_b$ besides its parameter settings and the corresponding forward input feature maps $u$. A general step of the backward process is: Firstly, this algorithm screens an element index set $I_B$, in which all the corresponding elements meet the condition that the input element should be less than or equal to 0, and the inherited element of gradients should be greater than 0. Then it calculates the backward value of the corresponding elements in the set $I_B$, and selects a part of the most valuable elements by sort rate $s$ to compose the index collection $G$. Finally, we modify the derivative method of ReLU at the elements corresponding to the index set $G$, to reset the gradients of these elements so that improve the ``wrong blocking".

\subsection{Improvement for ``over transmission"}
In order to alleviate or even eliminate the phenomenon of ``over transmission", and meanwhile avoid to cause excessive perturbation, we can screen the elements which are severely ``over transmission". Therefore, similar to the ADV-ReLU-B, we adopt the same solution that mapping the selected value of elements to the real number field, to sort and screen. And we set these corresponding derivatives of ReLU at these elements to 0 to prevent this problem.

Similar to ADV-ReLU-B, we also propose a new algorithm (called ADV-ReLU-T) to address this issue, as shown in Algorithm 2. Nevertheless, there are three differences between them: (1) Because the principle of ``over transmission" is different from the other, this algorithm selects the elements greater than 0 and its inherited gradient is less than 0. (2) In the calculation of the collection set $V$, it is extra multiplied by $-1$ so that it also can be in the descending sorting. (3) It sets the elements of derivative of ReLU: $g_{\sigma}$ at index collection $G$ should be $0$ rather than $1$, which therefore alleviates ``over transmission".

\begin{algorithm}[t] 
	\caption{ADV-ReLU-T} 
	\label{alg.ADV-ReLU-after} 
\renewcommand{\algorithmicrequire}{\textbf{Input:}}  
\renewcommand{\algorithmicensure}{\textbf{Output:}} 
	\begin{algorithmic}[1] 
		\Require Input feature maps $u$ for ReLU, inherited gradient $g_{b}=\frac{\partial J(u)}{\partial \sigma(u)}$, sort rate $s$, and a constant $c$.
		\Ensure Modified gradient $g=\frac{\partial{J(u)}}{\partial u}$.
		\State{Select and obtain the set $I_T = \{ i | u[i] > 0, g_b[i]<0\}$; \!\!\hfill $/\!/$\,\emph{Similar to Algorithm \ref{alg.ADV-ReLU-before} at Step 1}}
		\State Calculate the set $V=\left\{-\left(c \cdot u[i]+g_{b}[i]\right)|i \in I_T\right\}$;
		\State Sort $V$ in a descending order;
		\State Select $\lfloor s \cdot \left| V \right|\rfloor$ largest elements in $V$ to obtain $G$, where the number is $\left| G \right| = \lfloor s \cdot \left| V \right|\rfloor$;
		\State Let $g_{\sigma}=\frac{\partial \sigma(u)}{\partial u}$, then cover the elements at the index set $G$ to stop the redundant gradients, i.e., $g_{\sigma}[G]=0$;
		\State $g=g_{b} \odot g_{\sigma}$;
		\State return  $g$.
	\end{algorithmic}
\end{algorithm}
\subsection{Our ADV-ReLU Algorithm }
We present a straightforward algorithm, called ADV-ReLU, to alleviate both issues of ``wrong blocking" and ``over transmission", as shown in Algorithm 3,  which combines the above two algorithms in their action scope of input feature map $x$. We show that it can help the normal ReLU obtain more guiding gradients during backward prorogation.

As there exists no non-overlapping in key processing data between ADV-ReLU-B and ADV-ReLU-T, we can simply input the two kind of indexes satisfying two conditions into the two algorithms respectively. Then we make the rest of elements to backward according to the original ReLU and output the combined results from the two improved algorithms.

However, it is important to notice that when we apply ADV-ReLU to obtain gradients of a loss function, the gradient can only stand for a more accurate direction in the increase of the loss function. In other words, the negative of this gradient \textbf{is not} a more accurate direction toward the decrease of the loss function. For the deep neural network that applied ADV-ReLU to backward, we can describe this property that $\frac{\partial (-loss)}{\partial x}$ \textbf{is not equal to} $-1\cdot\frac{\partial loss}{\partial x}$.

\renewcommand{\algorithmicrequire}{\textbf{Input:}}  
\renewcommand{\algorithmicensure}{\textbf{Output:}} 
\begin{algorithm}[!ht] 
	\caption{ADV-ReLU} 
	\label{alg3} 
	\begin{algorithmic}[1] 
		\Require Input feature maps $u$ for ReLU, the inherited gradient $g_{b}=\frac{\partial J(u)}{\partial \sigma(u)}$, sort rate $s$, and a constant $c$.
		\Ensure Modified gradient $g=\frac{\partial{J(u)}}{\partial u}$.
		\State Select and obtain the set $I_B = \{ i | u[i]\le0,{g_b}[i]>0\}$;
		\State Select and obtain the set $I_T = \{ i | u[i] > 0, g_b[i]<0\}$;
		\State $g=\frac{\partial \sigma(u)}{\partial u} \odot g_{b}$;
		\State $g\left[I_B\right]=\operatorname{ADV-ReLU-B}\left(u\left[I_B\right], g_{b}\left[I_B\right], s, c\right)$;
		\State $g\left[I_T\right]=\operatorname{ADV-ReLU-T}\left(u\left[I_T\right], g_{b}\left[I_T\right], s, c\right)$;
		\State return  $g$.
	\end{algorithmic}
\end{algorithm}

\begin{table*}[htbp]
\centering
\caption{The perturbation (mean of the $\ell_2$ norm) and its reduction rate of generating adversarial examples without/with ADV-ReLU under white-box attacks. An asterisk * indicates that the algorithm is with our ADV-ReLU, and $\triangle$ denotes reduced perturbation with our ADV-ReLU.}
\label{tab.whitebox}
\setlength{\tabcolsep}{5.6pt}
\renewcommand\arraystretch{1.5}
\begin{tabular}{c|cccccccccc}
\hline
         Attack methods & VGG16-bn & $\triangle$ & Resnet101 & $\triangle$ & Nasnetamobile &$\triangle$ & Inception-v3 & $\triangle$ & Adv-inception-v3 & $\triangle$  \\
        \hline\hline
         FGSM & 3.6624 & & 6.0473 & & 16.2541 & & 26.6018 & & 101.4879 &  \\
         FGSM* & \textbf{3.5979} &1.8\%& \textbf{5.5843} &7.7\%& \textbf{13.2168} &18.7\%& \textbf{22.2562} &16.3\%& \textbf{101.2253} &0.3\% \\
        \hline
         I-FGSM & 0.5383 & & 0.6491 & & 0.7550 & & 1.0255 & & 0.9800 & \\
         I-FGSM* & \textbf{0.5329} &1.0\%& \textbf{0.6383} &1.7\%& \textbf{0.7027} &6.9\%& \textbf{0.9862} &3.8\%& \textbf{0.9364} &4.4\% \\
        \hline
         MI-FGSM & 0.5766 & & 0.7091 & & 0.8615 & & 1.1402 & & 1.1751 & \\
         MI-FGSM* & \textbf{0.5688} &1.4\%& \textbf{0.6976} &1.6\%& \textbf{0.7933} &7.9\%& \textbf{1.0882} &4.6\%& \textbf{1.1180} &4.9\% \\
        \hline
         C\&W & \textbf{0.3698} & & \textbf{0.5284} & & 0.5250 & & 0.8071 & & 0.5820 &  \\
         C\&W* & 0.3737 &-1.1\%& 0.5354 &-1.3\%& \textbf{0.5085} &3.1\%& \textbf{0.7723} &4.3\%& \textbf{0.5592} &3.9\% \\
        \hline
         Curls\&Whey & 0.2170 & & \textbf{0.3567} & & 0.4087 & & 0.6012 & & 0.6251 &  \\
         Curls\&Whey* & \textbf{0.2150} &0.9\%& 0.3568 &0.0\%& \textbf{0.3884} &5.0\%& \textbf{0.5662} &5.8\%& \textbf{0.5945} &4.9\% \\
        \hline
    \end{tabular}
\end{table*}

\begin{table*}[htbp]
	\centering
	\caption{The success rate s(\%) of using Inceptionresnet-v2 substitution without/with ADV-ReLU under black-box adversarial attacks. An asterisk * indicates that the adversarial attack method is with our ADV-ReLU.}\label{tab.blackbox2}
\setlength{\tabcolsep}{15.6pt}
	\renewcommand\arraystretch{1.5}
	\begin{tabular}{c|ccccc}
		\hline
		Attack methods & Inception-v3 & Inception-v4 & Inceptionresnet-v2 & Xception & Adv-inception-v3 \\
		\hline\hline
		I-FGSM & 84.9\% & 77.7\% & 100.0\% & 81.8\% & 48.8\% \\
		I-FGSM* & \textbf{88.9\%} & \textbf{84.7\%} & 100.0\% & \textbf{85.3\%} & \textbf{48.9\%}  \\
		\hline
		MI-FGSM & 94.4\% & 90.3\% & 100.0\% & 92.0\% & 69.2\% \\
		MI-FGSM* & \textbf{96.7\%} & \textbf{95.3\%} & 100.0\% & \textbf{96.0\%} & \textbf{74.1\%} \\
		\hline
	\end{tabular}
\end{table*}

The reason for this property is obviously that the directions of  input inherited gradients in these two expressions are not equal. And we have made assumptions with $\mathop {\max }\limits_{x'} {\rm{ }}J(x')$ for the derivation of our algorithm in Section \ref {TwoDrawbacks}. So when applying our algorithms, we can convert the minimizing problem of the loss function into its equivalent maximizing one. That is, we convert $\mathop {\min }\limits_x {\rm{ }}loss$ into $\mathop {\max }\limits_x {\rm{ }}(-loss)$.

\begin{figure}[h]
\centering
\includegraphics[width=1.006\columnwidth]{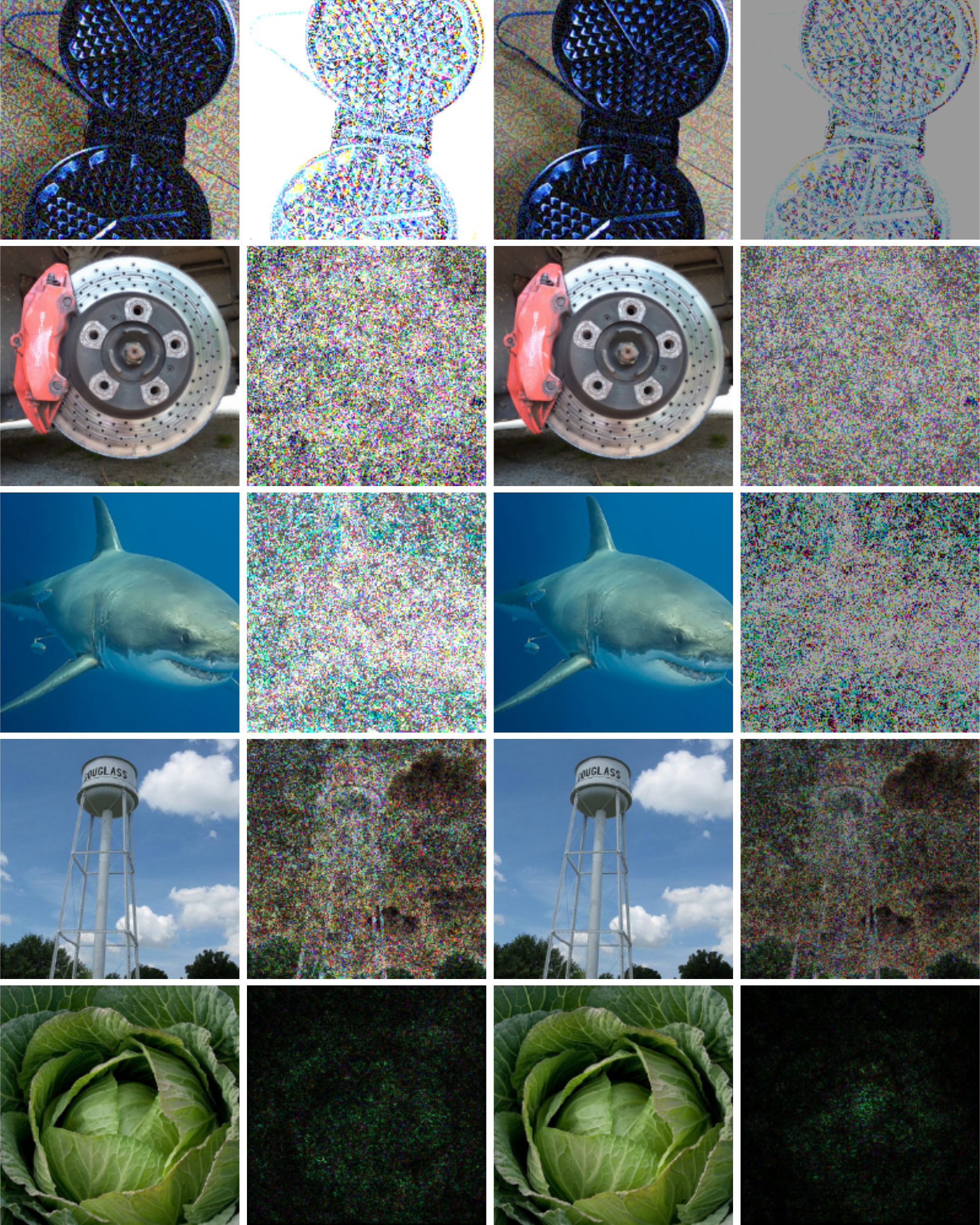}
\caption{The adversarial examples and their perturbations generated by the five state-of-the-art algorithms: FGSM (1st row), I-FGSM (2nd row), MI-FGSM (3rd row), C\&W (4th row), and Curls\&Whey (5th row), and their counterparts integrated with our ADV-ReLU.}
\label{figExamps}
\end{figure}

\section{Experiments}
In this section, we integrate our ADV-ReLU algorithm into some typical gradient-based white-box attack algorithms and attack some classical network structures on the ImageNet data set \cite{Russakovsky2015ImageNet} to evaluate the attack efficiency. Moreover, we also test our algorithm for black-box attacks to show its transferability for different models.

Below we first describe the parameter settings. Then we give the perturbation efficiency under integrated white-box attacks. We also report the results of success rate under integrated black-box attacks. Finally, we show the influence of parameter settings on attack efficiency.

\subsection{Experimental Setup}
The Imagenet2012 verification set is used as benchmark for the evaluation, and we implement the following neural networks: Resnet101 \cite{He2016DeepRL}, Vgg19-bn \cite{Simonyan2014Very}, NASNetAMobile \cite{Zoph2018LearningTA}, Inception-v3 \cite{szegedy2016rethinking}, Inception-v4 \cite{Szegedy2016Inception}, Inceptionresnet-v2 \cite{Szegedy2016Inception}, Xception \cite{Chollet2017XceptionDL} and adv-inception-v3 \cite{Kurakin2017AdversarialML} (adv-Inception-v3 is a robust model applied adversarial training). We select 500 categories, 2 images in each category from the Imagenet2012 verification set, and all these 1,000 images can be completely correctly classified by the above models.

Some classic algorithms for generating adversarial examples, such as FGSM \cite{goodfellow2014explaining}, I-FGSM \cite{kurakin2016adversarial}, MI-FGSM \cite{dong2018boosting}, C\&W \cite{Carlini2016Towards}, and Curls\&Whey \cite{Shi2019CurlsW}, are used to test, and we compare them with ADV-ReLU. For their implementations, we refer to the Foolbox (version 2.3.0) \cite{Rauber2017FoolboxVA}. For the implementation of Curls\&Whey \cite{Shi2019CurlsW}, we used the authors' source code. In addition, in order to match the optimization form of ADV-ReLU and loss function, we convert the minimization problem of the loss function to its equivalent maximization problem.

Moreover, the success rate and norm distance are used to quantitatively evaluate the effectiveness of the attack algorithms. Obviously, the success rate is the ratio that the number of successfully attacked examples divided by the total adversarial examples. The norm distance is defined as the average distance in the $l_2$ norm as follows:
\begin{equation}
  d(N) = \frac{1}{{|N|}}\sum\limits_{{x_i} \in N} {{{\left\| {{x_i} - x_i^\prime } \right\|}_2}}
\end{equation}
where $N$ represents the set of  examples, and $|N|$ represents the number of  examples of the set $N$. $x_i$ is the example with index $i$ in the set $N$, and $x'_i$ is the corresponding adversarial example at the same index.

\begin{figure*}[htb]
	\centering  
	\subfigure[General trends]{
		\includegraphics[width=0.45\textwidth,height=0.35\textwidth]{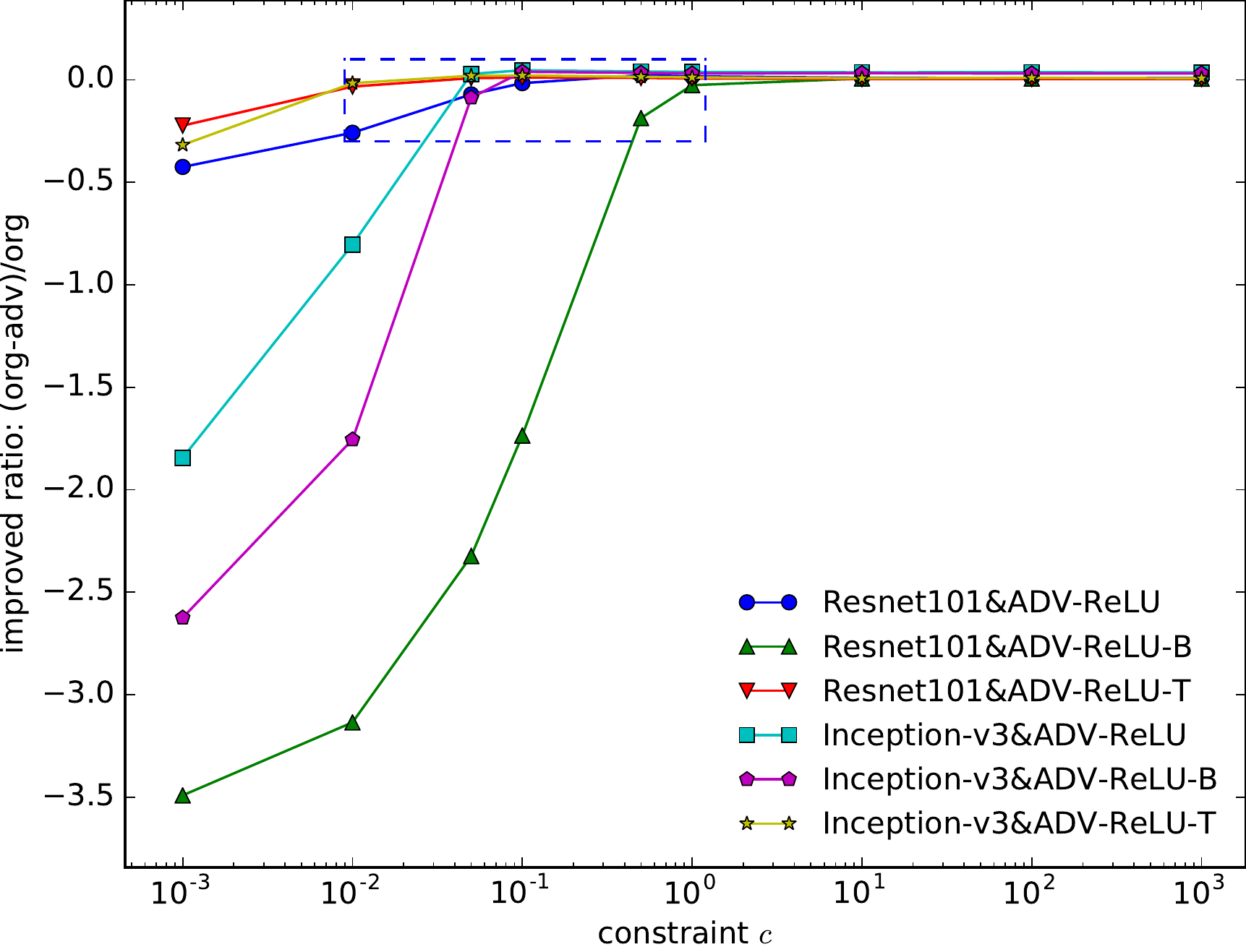}}
	\subfigure[Local trends in dotted area]{
		\includegraphics[width=0.45\textwidth,height=0.35\textwidth]{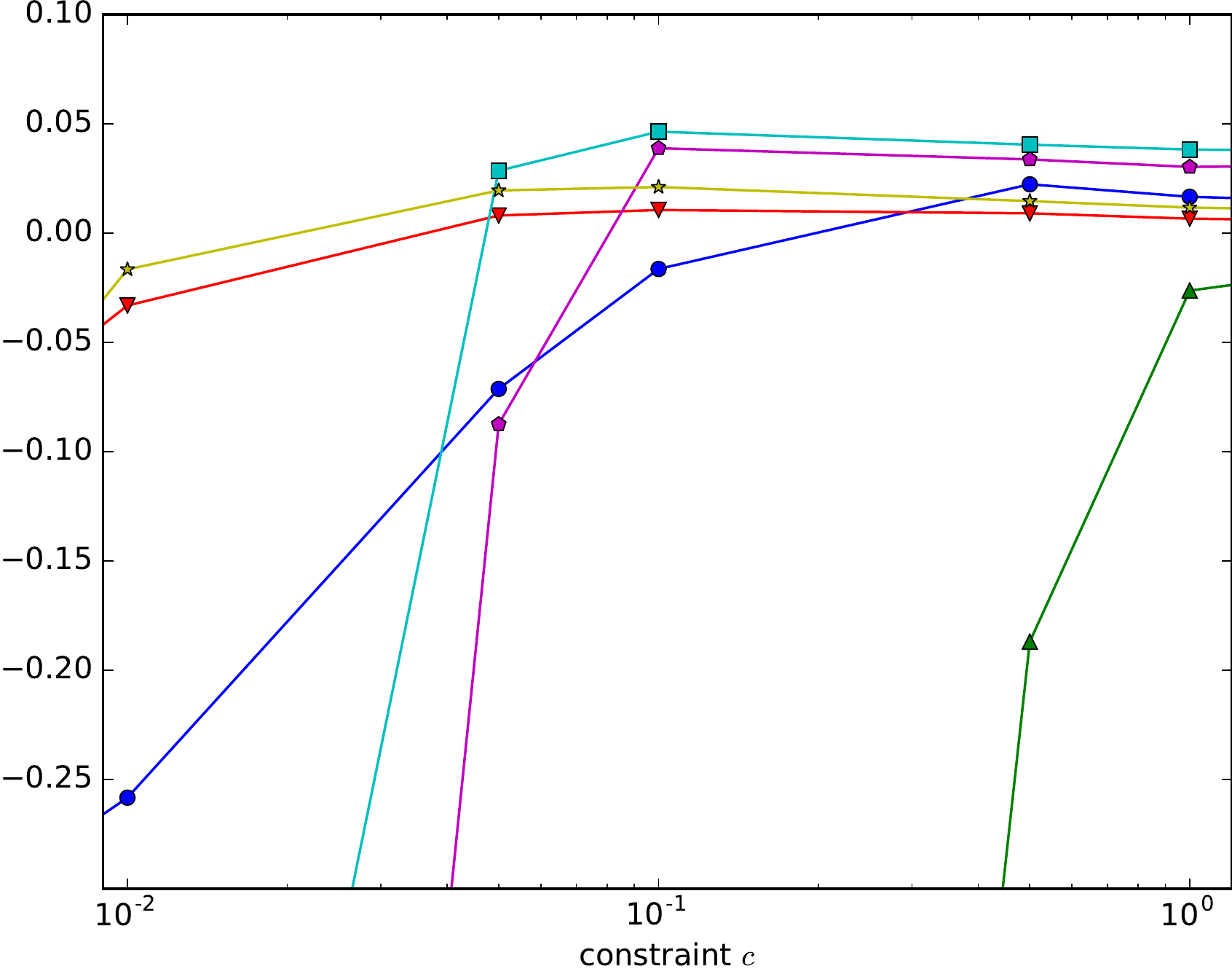}}
	\caption{The reduced perturbations using ADV-ReLU and the sub-algorithms: ADV-ReLU-B and ADV-ReLU-T vary with the parameter constraint $c$. (a) global view of the variation tendency (b) zoomed local view near the peaks. We can see that there exist much consistency even in different networks.}\label{Fig.constraint}	
\end{figure*}
\begin{figure*}[htb]
	\centering  
	\subfigure[General trends]{
		\includegraphics[width=0.45\textwidth,height=0.35\textwidth]{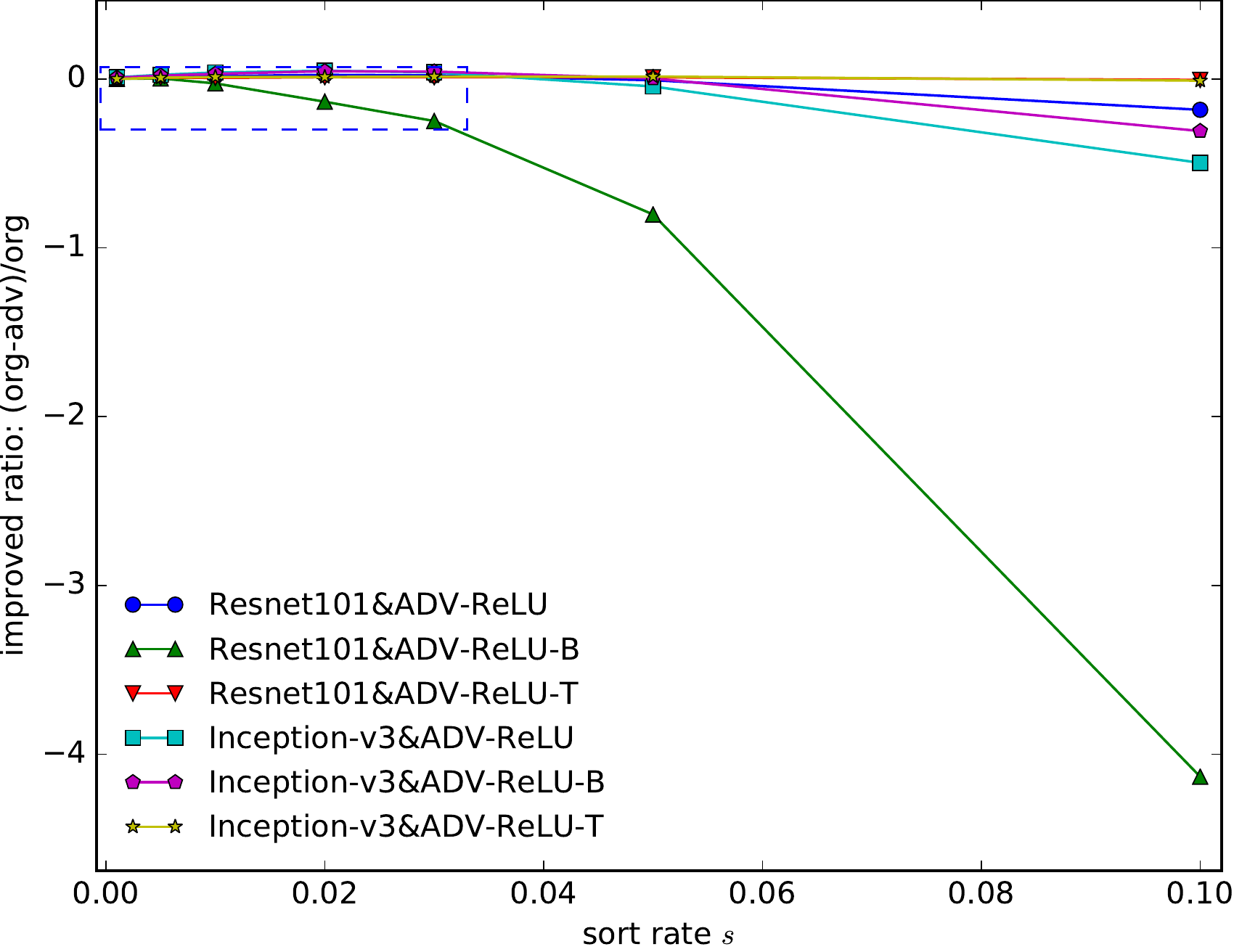}}
	\subfigure[Local trends]{
		\includegraphics[width=0.45\textwidth,height=0.35\textwidth]{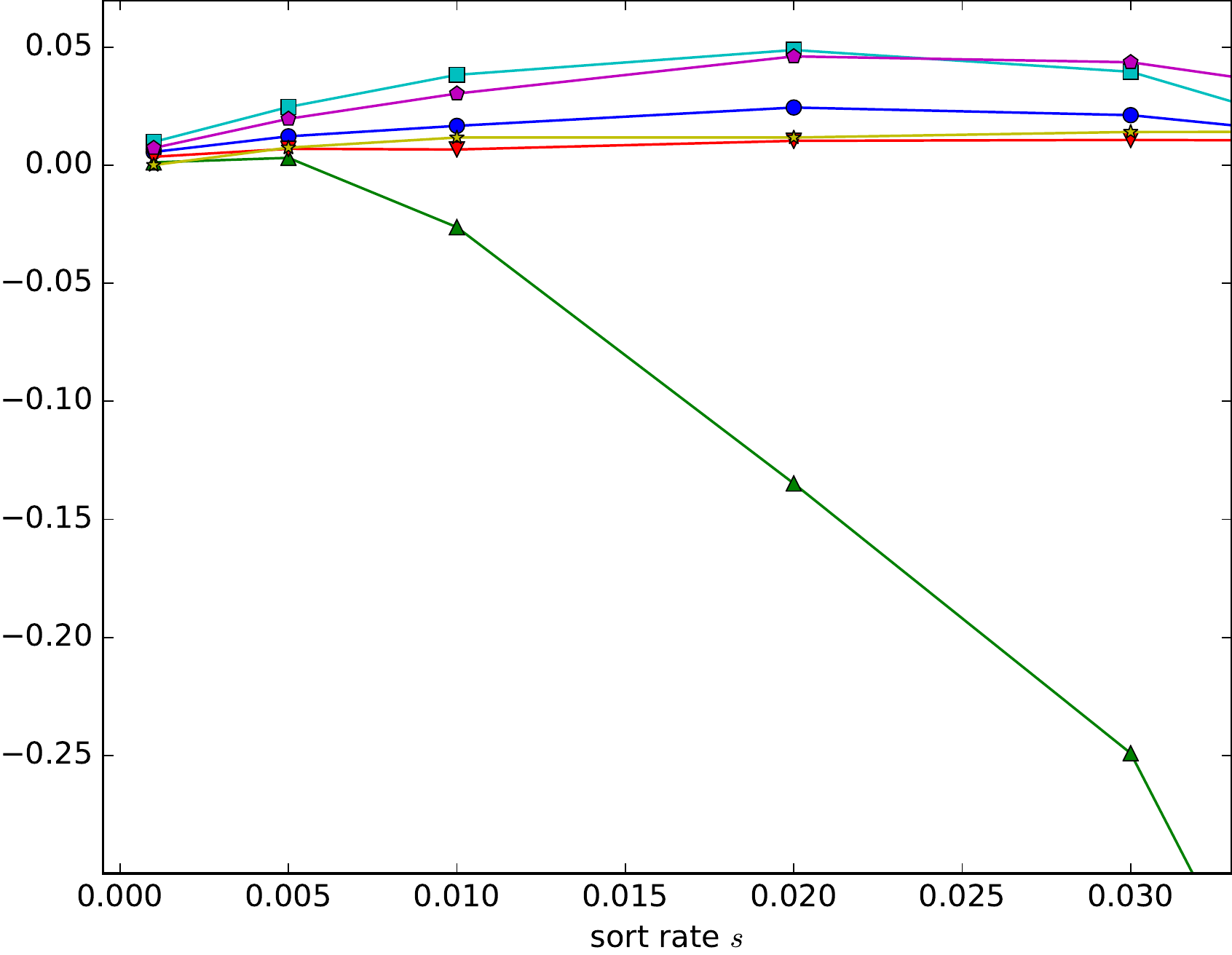}}
	\caption{The reduced perturbations using ADV-ReLU and the sub-algorithms (i.e., ADV-ReLU-B and ADV-ReLU-T) vary with the parameter sort rate $s$. (a) Global view of the variation tendency, and (b) zoomed local view near the peaks.}
	\label{Fig.sortrate}	
\end{figure*}
\subsection{Experiments on typical white-box attacks}
For the FGSM algorithm, since a fixed step size hardly shows the difference using the original ReLU and ADV-ReLU, we search their suitable step size to show their performance. For the remaining four iteratively adversarial attack algorithms, we uniformly set the maximum perturbation size to $\varepsilon=0.1$ (in the ${\ell _\infty}$ norm, and the range of image pixel value is [0,1]), step size as $\alpha=0.001$, which allow the algorithm to search for sufficiently small adversarial examples, and the maximum number of iterations as 100. Other parameters are set as follows. In MI-FGSM, decay factor is set to $\mu=0.5$. In Curls\&Whey,  the binary search number is 3. And for the white-box attacks integrating our ADV-ReLU algorithm, we set $s=0.01$, $c=1$. Except for the C\&W attacker with Resnet101, the sort rate is set to $s=0.001$ for obtaining a superior result.

In Table \ref{tab.whitebox}, the perturbations in the mean $\ell_2$ norm of generated adversarial examples for each attack algorithm and the integrated one with ADV-ReLU, are shown. It indicates that the performance of most white-box algorithms integrated with ADV-ReLU can be improved, as they gain reduced perturbations compared with that of the original algorithms. FGSM integrated with ADV-ReLU attacking NASNetAMobile attains an perturbation of 13.2168, which is reduced by 18.7\% compared with that of the original FGSM. And when FGSM with ADV-ReLU attacks Inception-v3, it reduces about 16.3\% perturbations compared with the original one. This may result from the wider distribution of the weights in the networks, which leads to more obvious effect.  Moreover, it is noted that when C\&W attacks VGG16-bn and Resnet101 networks, the perturbations with ADV-ReLU are slightly larger than those of the original ones. It is probably because that ``wrong blocking" and ``over transmission" occur more severely in the backpropagation in these two networks.

Moreover, the generated adversarial examples by the five state-of-the-art algorithms: FGSM, I-FGSM, MI-FGSM, C\&W, and Curls\&Whey, attacking the Inception-v3 network and their integrated variants with our ADV-ReLU are shown in Fig.\ \ref{figExamps}. The perturbations are magnified to highlight the differences as in \cite{Zhang2020PatchWise}. It can be seen that the algorithms with ADV-ReLU yield darker perturbations in the fourth column than those of the original white-box attack algorithms in the second column, though there seems no obvious differences between the generated adversarial examples. All the results indicate that the proposed ADV-ReLU algorithm can produce superior adversarial examples and can be used to those popular white-box attack algorithms.
\begin{figure}[htb]
	\centering  
	\includegraphics[width=0.45\textwidth,height=0.35\textwidth]{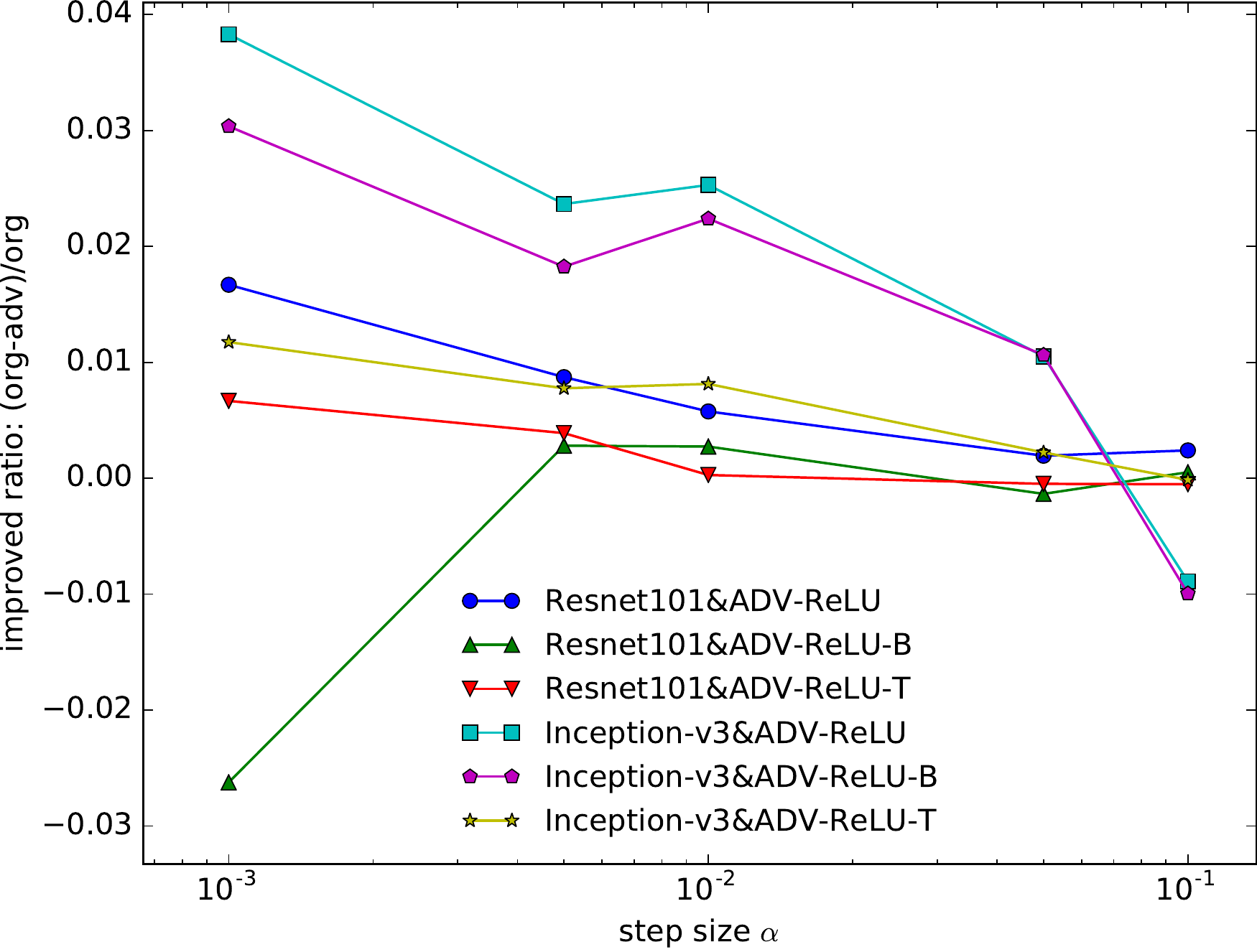}
	\caption{The reduced perturbations using ADV-ReLU and the sub-algorithms (i.e., ADV-ReLU-B and ADV-ReLU-T) vary with the parameter step size $\alpha$. (a) Global view of the variation tendency, and (b) zoomed local view near the peaks.}\label{Fig.stepsize}	
\end{figure}

\subsection{Experiments on typical black-box attacks}
In order to show the transferability of the proposed framework, we also compare our algorithm against the black-box non-target attacks. Here, the attack methods are  I-FGSM and MI-FGSM, and the substitute model is Inceptionresnet-v2, whose gradient is used for the attack methods to generate adversarial examples. The target networks include Inception-v3, Inception-v4, Inceptionresnet-v2, Xception, and adv-inception-v3. The parameters for I-FGSM and MI-FGSM are set as follows: maximum perturbation size $\varepsilon=0.3$, step size $\alpha=0.01$, and $iterations=100$. And the decay factor is set to $\mu=0.5$ for MI-FGSM. The attacking success rates of all the algorithms are listed in Table \ref{tab.blackbox2}.

From Table \ref{tab.blackbox2}, it can be seen that when Inceptionresnet-v2 is used as an substitution model, all the success rates of the I-FGSM and MI-FGSM attack algorithms increase. Especially, I-FGSM with ADV-ReLU attacking Inception-v4 gains an improvement of 7.0\% success rate compared with that of I-FGSM. The reason could be that the gradient distribution of Inceptionresnet-v2 are more non-smooth than the above Inception series networks, leading to a decrease in transferability. Because AVD-ReLU can provide a more precise gradient and alleviate the non-smooth phenomenon, and its improves the consistency of Inceptionresnet-v2 across other Inception series networks.

\subsection{Parameter discussion}
In this subsection, the performance of our ADV-ReLU algorithm is studied under different parameter settings: a constraint constant $c$, a sort rate $s$ and a step size $\alpha$, Note that each time only one parameter varies, and the others remain as in those typical white-box attacks. The white-box non-target attack, I-FGSM, is against two networks: Resnet101 and Inceptionv3, respectively. And the decreased perturbations of the attack algorithm using ADV-ReLU (in ratio) respect to the original algorithm are shown in Figs.\  \ref{Fig.constraint} and \ref{Fig.sortrate}. Moreover, in order to show the influence of the parameters to the two sub-algorithms (i.e., ADV-ReLU-B and ADV-ReLU-T), we also plotted them in the figures.

\subsubsection{Parameter $c$}
As shown in Fig.\ \ref{Fig.constraint}, the constant parameter $c$ influences the perturbations. It plays an important role in evaluating the value of the elements. It balances the input value of ReLU and the inherited gradient, so that elements selected by ADV-ReLU can relieve ``wrong blocking" and ``over transmission" after derivation. If this parameter is too large, ADV-ReLU tends to select the indexes at where absolute value of the elements are very close to $0$. If this parameter is too small, ADV-ReLU tends to select the indexes at where absolute value of inherited gradients are very large. From Fig.\ \ref{Fig.constraint}, we can see that, except Resnet101 under ADV-ReLU-B, as $c$ increases, the perturbation gradually rises. When $c>0.1$, the perturbation tends to stable. This indicates that when $c$ is lager than 0.1, the performance of ADV-ReLU is superior.

\subsubsection{Sort rate $s$}
The decreased perturbations of the attack algorithm using ADV-ReLU (in ratio) and its sub-algorithms are shown in Fig.\ \ref{Fig.sortrate}. The sort rate $s$ indicates how many elements are selected from the the feature map during the ADV-ReLU backpropagation. As the parameter increases from 0.001 to 0.10, the perturbation of the attack algorithm decreases gradually. An optimal range is within $[0.01,0.03]$. It is worth noting that when ADV-ReLU-B is integrated and attack Resnet101, sort rate over $0.005$ can dramatically reduce the performance. This is probably because the distributions of the values in feature maps in each layer of the network with residual layer is smaller than that in other network architectures, and a smaller sort rate is more suitable for the derivation.

\subsubsection{Step size $\alpha$}
The decreased perturbations of the attack algorithm using ADV-ReLU (in ratio) and its sub-algorithms are shown in Fig.\ \ref{Fig.stepsize}.
The step size always influences the level of the effective perturbation. Our algorithms with small step sizes gain larger levels of perturbations than those of large step sizes. This is in accordance with the analysis and discussion in \cite{Shi2019CurlsW}. The smaller the step size and the greater the number of iterations, the closer the path that generates the perturbation is to a smooth gradient curve. That is, the advantage of ADV-ReLU is more evident when the improved gradient generates smooth paths, and fits better with the guidance of the gradient. Therefore, in the experiments, we set step size as 0.001.

\section{Conclusions}
In this paper, we proposed a novel universal adversarial example generation framework, called ADV-ReLU, which can be easily integrated into the gradient-based white-box attack algorithms. The key idea of the algorithm is to improve the two defects of the ReLU activation function in the process of backpropagation for training, so that the calculated gradients of the improved network are more consistent with the actual changes. The experimental results on ImageNet show that our framework can be embedded with many powerful white-box attack algorithms and further improve the performance of all the algorithms. And the experiments on black-box attacks indicate that our framework enjoys sound transferability for various networks. Therefore, we believe that our framework can be also used to train more robust networks. In the future, we will figure out more strategies to re-assign gradients to improve the performance of our algorithm.

\section*{Acknowledgment}
We thank all the reviewers for their valuable comments. 

\bibliographystyle{IEEEtran}
\bibliography{ref}

\begin{thebibliography}{10}
\providecommand{\url}[1]{#1}
\csname url@samestyle\endcsname
\providecommand{\newblock}{\relax}
\providecommand{\bibinfo}[2]{#2}
\providecommand{\BIBentrySTDinterwordspacing}{\spaceskip=0pt\relax}
\providecommand{\BIBentryALTinterwordstretchfactor}{4}
\providecommand{\BIBentryALTinterwordspacing}{\spaceskip=\fontdimen2\font plus
\BIBentryALTinterwordstretchfactor\fontdimen3\font minus
  \fontdimen4\font\relax}
\providecommand{\BIBforeignlanguage}[2]{{%
\expandafter\ifx\csname l@#1\endcsname\relax
\typeout{** WARNING: IEEEtran.bst: No hyphenation pattern has been}%
\typeout{** loaded for the language `#1'. Using the pattern for}%
\typeout{** the default language instead.}%
\else
\language=\csname l@#1\endcsname
\fi
#2}}
\providecommand{\BIBdecl}{\relax}
\BIBdecl

\bibitem{szegedy2013intriguing}
C.~Szegedy, W.~Zaremba, I.~Sutskever, J.~Bruna, D.~Erhan, I.~Goodfellow, and
  R.~Fergus, ``Intriguing properties of neural networks,'' in \emph{Proc. Int.
  Conf. Learn. Represent. (ICLR)}, 2014.

\bibitem{goodfellow2014explaining}
I.~J. Goodfellow, J.~Shlens, and C.~Szegedy, ``Explaining and harnessing
  adversarial examples,'' in \emph{Proc. Int. Conf. Learn. Represent. (ICLR)},
  2015.

\bibitem{kurakin2016adversarial}
A.~Kurakin, I.~J. Goodfellow, and S.~Bengio, ``Adversarial examples in the
  physical world,'' in \emph{Proc. Int. Conf. Learn. Represent. (ICLR)}, 2019.

\bibitem{Carlini2016Towards}
N.~Carlini and D.~Wagner, ``Towards evaluating the robustness of neural
  networks,'' in \emph{Proc. IEEE Symp. Secur. Privacy (SP)}, vol.~0, 2017, pp.
  39 -- 57.

\bibitem{MoosaviDezfooli2016DeepFoolAS}
S.-M. Moosavi-Dezfooli, A.~Fawzi, and P.~Frossard, ``Deepfool: A simple and
  accurate method to fool deep neural networks,'' in \emph{Proc. IEEE Comput.
  Soc. Conf. Comput. Vision Pattern Recognit. (CVPR)}, vol. 2016-December,
  2016, pp. 2574 -- 2582.

\bibitem{dong2018boosting}
Y.~Dong, F.~Liao, T.~Pang, H.~Su, J.~Zhu, X.~Hu, and J.~Li, ``Boosting
  adversarial attacks with momentum,'' in \emph{Proc. IEEE Comput. Soc. Conf.
  Comput. Vision Pattern Recognit. (CVPR)}, 2018, pp. 9185 -- 9193.

\bibitem{wu2018understanding}
\BIBentryALTinterwordspacing
L.~Wu, Z.~Zhu, C.~Tai, and E.~Weinan, ``Understanding and enhancing the
  transferability of adversarial examples,'' 2018, \emph{arXiv:1802.09707v1}.
  [Online]. Available: \url{https://arxiv.org/abs/1802.09707v1}
\BIBentrySTDinterwordspacing

\bibitem{moosavi2017universal}
S.-M. Moosavi-Dezfooli, A.~Fawzi, O.~Fawzi, and P.~Frossard, ``Universal
  adversarial perturbations,'' in \emph{Proc. IEEE Conf. Comput. Vis. Pattern
  Recognit. (CVPR)}, vol. 2017-January, 2017, pp. 86 -- 94.

\bibitem{Shi2019CurlsW}
Y.~Shi, S.~Wang, and Y.~Han, ``Curls whey: Boosting black-box adversarial
  attacks,'' in \emph{Proc. IEEE Comput. Soc. Conf. Comput. Vis. Pattern
  Recognit. (CVPR)}, vol. 2019-June, 2019, pp. 6512 -- 6520.

\bibitem{Tram2017Ensemble}
F.~Tramer, A.~Kurakin, N.~Papernot, I.~Goodfellow, D.~Boneh, and P.~McDaniel,
  ``Ensemble adversarial training: Attacks and defenses,'' in \emph{Proc. Int.
  Conf. Learn. Represent. (ICLR)}, 2018.

\bibitem{Xie2019FeatureDF}
C.~Xie, Y.~Wu, L.~V.~D. Maaten, A.~L. Yuille, and K.~He, ``Feature denoising
  for improving adversarial robustness,'' in \emph{Proc. IEEE Comput. Soc.
  Conf. Comput. Vis. Pattern Recognit. (CVPR)}, vol. 2019-June, 2019, pp. 501
  -- 509.

\bibitem{AdSurvey19}
X.~{Yuan}, P.~{He}, Q.~{Zhu}, and X.~{Li}, ``Adversarial examples: Attacks and
  defenses for deep learning,'' \emph{IEEE Trans. Neural Networks Learn. Sys.
  (TNNLS)}, vol.~30, no.~9, pp. 2805--2824, 2019.

\bibitem{Ilyas2019Adversarial}
A.~Ilyas, S.~Santurkar, D.~Tsipras, L.~Engstrom, B.~Tran, and A.~Madry,
  ``Adversarial examples are not bugs, they are features,'' in \emph{Proc. Adv.
  Neural Inf. Process. Syst. (NIPS)}, 2019.

\bibitem{AdSurvey20}
J.~{Zhang} and C.~{Li}, ``Adversarial examples: Opportunities and challenges,''
  \emph{IEEE Trans. Neural Networks Learn. Sys. (TNNLS)}, vol.~31, no.~7, pp.
  2578--2593, 2020.

\bibitem{tnnlsGrad}
Z.~{Katzir} and Y.~{Elovici}, ``Gradients cannot be tamed: Behind the
  impossible paradox of blocking targeted adversarial attacks,'' \emph{IEEE
  Trans. Neural Networks Learn. Sys. (TNNLS)}, pp. 1--11, 2020.

\bibitem{Papernot2016TransferabilityIM}
\BIBentryALTinterwordspacing
N.~Papernot, P.~McDaniel, and I.~J. Goodfellow, ``Transferability in machine
  learning: from phenomena to black-box attacks using adversarial samples,''
  2016, \emph{arXiv:1605.07277}. [Online]. Available:
  \url{https://arxiv.org/abs/1605.07277}
\BIBentrySTDinterwordspacing

\bibitem{Clevert2015FastAA}
D.-A. Clevert, T.~Unterthiner, and S.~Hochreiter, ``Fast and accurate deep
  network learning by exponential linear units (elus),'' in \emph{Proc. Int.
  Conf. Learn. Represent. (ICLR)}, 2016.

\bibitem{Maas13rectifiernonlinearities}
A.~L. Maas, A.~Y. Hannun, and A.~Y. Ng, ``Rectifier nonlinearities improve
  neural network acoustic models,'' in \emph{Proc. Int. Conf. Mach. Learn.
  (ICML)}, 2013.

\bibitem{Russakovsky2015ImageNet}
O.~Russakovsky, J.~Deng, H.~Su, J.~Krause, S.~Satheesh, S.~Ma, Z.~Huang,
  A.~Karpathy, A.~Khosla, M.~Bernstein, A.~C. Berg, and L.~Fei-Fei, ``Imagenet
  large scale visual recognition challenge,'' \emph{Int. J. Comput. Vision
  (IJCV)}, vol. 115, no.~3, pp. 211 -- 252, 2015.

\bibitem{He2016DeepRL}
K.~He, X.~Zhang, S.~Ren, and J.~Sun, ``Deep residual learning for image
  recognition,'' in \emph{Proc. IEEE Comput. Soc. Conf. Comput. Vis. Pattern
  Recognit. (CVPR)}, vol. 2016-December, 2016, pp. 770 -- 778.

\bibitem{Simonyan2014Very}
K.~Simonyan and A.~Zisserman, ``Very deep convolutional networks for
  large-scale image recognition,'' in \emph{Proc. Int. Conf. Learn. Represent.
  (ICLR)}, 2015.

\bibitem{Zoph2018LearningTA}
B.~Zoph, V.~Vasudevan, J.~Shlens, and Q.~V. Le, ``Learning transferable
  architectures for scalable image recognition,'' in \emph{Proc. IEEE Comput.
  Soc. Conf. Comput. Vis. Pattern Recognit. (CVPR)}, 2018, pp. 8697 -- 8710.

\bibitem{szegedy2016rethinking}
C.~Szegedy, V.~Vanhoucke, S.~Ioffe, J.~Shlens, and Z.~Wojna, ``Rethinking the
  inception architecture for computer vision,'' in \emph{Proc. IEEE Comput.
  Soc. Conf. Comput. Vis. Pattern Recognit. (CVPR)}, vol. 2016-December, 2016,
  pp. 2818 -- 2826.

\bibitem{Szegedy2016Inception}
C.~Szegedy, S.~Ioffe, V.~Vanhoucke, and A.~A. Alemi, ``Inception-v4,
  inception-resnet and the impact of residual connections on learning,'' in
  \emph{Proc. AAAI Conf. Artif. Intell. (AAAI)}, 2017, pp. 4278 -- 4284.

\bibitem{Chollet2017XceptionDL}
F.~Chollet, ``Xception: Deep learning with depthwise separable convolutions,''
  in \emph{Proc. IEEE Conf. Comput. Vis. Pattern Recognit. (CVPR)}, vol.
  2017-January, 2017, pp. 1800 -- 1807.

\bibitem{Kurakin2017AdversarialML}
A.~Kurakin, I.~J. Goodfellow, and S.~Bengio, ``Adversarial machine learning at
  scale,'' in \emph{Proc. Int. Conf. Learn. Represent. (ICLR)}, 2017.

\bibitem{Rauber2017FoolboxVA}
\BIBentryALTinterwordspacing
J.~Rauber, W.~Brendel, and M.~Bethge, ``Foolbox v0.8.0: A python toolbox to
  benchmark the robustness of machine learning models,'' 2017,
  \emph{arXiv:1707.04131}. [Online]. Available:
  \url{https://arxiv.org/abs/1707.04131}
\BIBentrySTDinterwordspacing

\bibitem{Zhang2020PatchWise}
L.~Gao, Q.~Zhang, j.~Song, X.~Liu, and H.~Shen, ``Patch-wise attack for fooling
  deep neural network,'' in \emph{Proc. Eur. Conf. Comput. Vis. (ECCV)}, 2020.

\end{thebibliography}

\end{document}